\documentclass[a4paper,10pt]{article}

\usepackage{AISB99}
\usepackage[english]{babel}

\begin{document}

\title{ {\LARGE\bf Mixing representation levels: \\The hybrid approach
    to automatic text generation } }

\author{
\Large
E. Pianta ;
L.M. Tovena ;
\\
\large ITC-IRST ;
\\via Sommarive
\\38050 Povo  TRENTO - Italy ;
\\
\large \{tovena,pianta\}@itc.it
}

\abstract{ Natural language generation systems (NLG) map non-linguistic
representations into strings of words through a number of steps using
intermediate representations of various levels of abstraction.
Template based systems, by contrast, tend to use only one
representation level, i.e.\ fixed strings, which are combined,
possibly in a sophisticated way, to generate the final text.
\\
In some circumstances, it may be profitable to combine NLG and template
based techniques.
The issue of combining generation techniques can be seen in more abstract
terms as the issue of mixing levels of representation of different
degrees of linguistic abstraction.  This paper 
aims at defining a reference architecture for systems using mixed
representations.
We argue that mixed representations can be used without abandoning
a linguistically grounded approach to language generation.
}

\maketitleandabstract

\section{Introduction}

Natural language generation systems (NLG) map non-linguistic
representations into strings of words through a number of steps using
intermediate representations of various levels of abstraction.
Template based systems, by contrast, tend to use only one
representation level, i.e.\ fixed strings, which are combined,
possibly in a sophisticated way, to generate the final text.

In some circumstances, it may be profitable to combine NLG and template
based techniques.
The issue of combining generation techniques can be seen in more abstract
terms as the issue of mixing levels of representation of different
degrees of linguistic abstraction.  This paper 
aims at defining a reference architecture for systems using mixed
representations.
We argue that using templates does not necessarily means abndoning
a linguistically grounded approach to language generation.

The rest of this paper is organised as follows: in section
\ref{NLGvsT} we introduce briefly the NLG and the template based approaches to text generation. Section
\ref{hybrid} offers a theoretical framework within which the hybrid approch combining these two strategies can be analysed. We also review
a number of existing systems from the point of view of the proposed
framework. Finally, section \ref{HTPL} presents a declarative formalism
that allows us to represent hybrid objects characterised
by different degrees of linguistic abstraction.

\section{NLG and templates \label{NLGvsT}}

The strategy to perform automatic text generation called Natural
Language Generation par excellence, or also Deep Generation, is
characterised by the fact that it relies on conceptual models of
language developed in current linguistic theories. Systems following
this strategy are based on concepts such as morpheme, word,
sentence, semantic or syntactic representation, communicative
intention, etc. Each of these objects pertains to a level of
linguistic analysis that has specific rules and specific ways to
represent information.

Current architectures of NLG systems are made of an ordered sequence
of components consisting of Text Planner, Sentence Planner and
Linguistic Realiser
\citep{Reiter94, ReiterDale97, Paiva98,
  CahillReape98}. Each component expects an input and produces an
output pertaining to a certain level of linguistic analysis.  More specifically:

\begin{itemize}
\item the input of the Text Planner is some information
represented in a non-linguistic formalism, for instance concepts of a
Knowledge Base. The pieces of information feeding the generation
process are usually called {\it Messages}.
\item  the output of the Text Planner and input of the Sentence
Planner is a {\it Text Plan}, that is a tree structure where the leaves
are Messages and the  nodes represent semantic relations
between subtrees which will be eventually realized as text spans. In
some case semantic
relations correspond to rhetorical relations.
\item  the output of the Sentence Planner and input of the
Linguistic Realiser is a list of {\it Sentence Representations}. A Sentence
Representation includes all the semantic and grammatical information
necessary to generate exactly one correct sentence according to the
syntactic, morphological and phonological rules of a certain
language.
\item  the ouput of the Linguistic Realiser is a list of strings
that constitute the generated text.

\end{itemize}

The NLG approach is usually compared with the template based one \citep{Reiter95}. The basic idea of a textual template is that of a structure
consisting of a sequence of fixed strings and gaps filled at
processing time with other fixed strings. Let's call it a template of
the {\em static} type. A static template can degenerate in canned text if no
gaps are present. Generation systems based solely on static templates
are called mail-merge systems. As suggested by \cite{ReiterDale97},
these systems can reach the complexity and flexibility of a
programming language, and are thus functionally equivalent to NLG
systems, at least in principle.

Pros and cons of both approaches have been discussed in the
literature, see \citep{ReiterMellish93, Reiter95, ReiterDale97,
  BusemannHoracek98}. Let us just mention some of them. 
\begin{itemize}
\item Pros of NLG:
declarativeness, theoretical soundness, modularity, good portability
through different domains i.e. reusability, aptness to handle
multilinguism. 
\item Cons of NLG: low time efficiency, architectural
complexity, linguistic resources are costly to develop and require
specialized knowledge, patches of poorly understood linguistic
phenomena, difficulties in integrating linguistic content and lay-out
information. 
\item Pros of static templates: high time efficiency,
architectural simplicity, efficient application development, only
generic programming skills are required. 
\item Cons of static templates: no
theoretical grounding, procedurality, low portability, multilinguism
is awkwardly handled.
\end{itemize}

In the last few years, templates have been used also within NLG
architectures, in order to overcome some of the drawbacks of NLG,
first of all time inefficiency and resource developing cost.
\cite{ReiterMellish93} introduce pointers to KB individuals among
fixed strings and insert canned text as the value of a frame
representing the meaning of a sentence. 
\cite{Busemann96} mixes templates  and syntactic representations.
\cite{Cancedda97} insert
templates as leaf nodes of a textual plan produced by classical NLG
techniques.  
These
attempts to integrate templates within NLG architectures all bring in
{\em mixed representations} (MR), absent from both pure NLG and
static templates. 

Our approach to text generation aimes at mixing representation levels
in a systematic and principled way. From a practical perspective, this
amounts to using precompiled generation knowledge whenever possible,
while retaining the possibility of using a full-fledged NLG approach
when strictly necessary.  

Let us see how the notion of mixed representation compares with the
received view about the separation of representation levels.  Standard
NLG architectures map an input message to an output text passing
through a number of intermediate representations, as we have seen.
Each representation level is the input and/or output of a separate
component coping with specific linguistic phenomena, e.g.
communicative intentions, text structure, referring expressions,
morphology, etc. In the accepted view, representation levels should be
kept carefully separated, on the grounds that separation enhances
modularity and reflects different levels of linguistic analysis. In
the Mixed Representation approach both these motivations are
challenged. 

On the one hand, we argue that representation levels can
be mixed while preserving the modularity of the linguistic components.
On the other hand, we argue that, while the strict separation of
representation levels is crucial when taking a competence point of
view on language, mixing representations is acceptable in a more
perfor\-man\-ce oriented perspective. In practical terms, we consider it
plausible that human speakers produce discourse by mixing dynamic
planning with precompiled knowledge about the structure and te relevance
of texts, and produce sentences by mixing flexible sentence planning
and realization with all kind of (semi)\-idiomatic expressions,
(semi)\-fixed descriptions of individuals, precompiled sentence
patterns, phrases stored in the short-term memory, etc.

Let us conclude this section by making a terminological point.
Instead of the opposition between NLG and templates introduced by
\cite{Reiter95}, more recently \cite{BusemannHoracek98} have proposed
a distinction between in-depth and shallow generation which parallels
the distinction between deep and shallow analysis. We think that this
proposal goes in the right direction for two reasons. First,
identifying natural language generation with deep generation seems
misleading and a little arbitrary. Second, as we will see in the rest
of the paper, templates are just one among various shallow generation
techniques available.

\section{The hybrid approach to automatic text generation \label{hybrid}}

In this section we try to characterize the so called hybrid approach
to text generation in terms of mixing representation levels. Before
doing that, we need to single out a few more levels of representation.

We start by
distinguishing three components which are virtually present in any
Linguistic Realiser, that is the Sentence Grammar, the Morphological
Synthesizer and the Phonological Adjustment Component. The little
attention paid so far to Morphological and Phonological components may
be explained by the fact that many of the generation systems described
in the literature produce texts in English, a language with relatively
little inflectional Morphology and a restricted number of Phonological
Adjustment phenomena. In principle, a Sentence Grammar can incorporate
the other two components, since it can produce directly complete words
and apply phonological adjustment rules as soon as a pair of adjacent
words is available. However, we would rather keep the three components
apart, in order to broaden the possible range of mixed
representations, as discussed below.
The distinctions introduced in the Linguistic Realiser implies new
representation levels.

\begin{itemize}
\item  the output of the Sentence Grammar and input of the
Morphological Synthesizer is a list of {\em morphological bundles}, i.e.\ 
sets of morphological features that are mapped onto potential words.
\item   the output of the Morphological Synthesizer and the input of
the Phonological Component is a list of {\em potential words}, which are
word forms that can undergo phonological adjustments.
\item   the output of the Phonological Component is a list of
strings.
\end{itemize}

We must spend a few words also on the task of {\it formatting}, which is
often underestimated in the literature.
More and more often generation systems are expected to produce
formatted text rather than bare ASCII code.  Formatting information is
taken in the broad sense of tags for typographical formatting,
pointers to images, hypertextual links, annotations for texts feeding
a speech synthesizer, etc. In order to be effective, formatting
decisions cannot be taken by a component independent from the NLG
architecture and subsequent to it. They need to be taken at early
stages of the NLG process, see \citep{ReiterMellishLevine95}. For instance, if you want
to emphasise typographically a certain portion of a sentence content,
you need to take this decision within the Sentence Planning component,
when the relevant semantic information is available. Or, if you want
your text to be articulated into sections and paragraphs, you need to
take decisions about these aspects during the Text Planning stage. The
actual execution of formatting instructions can be left to a component
that operates after the Linguistic Realiser, but the formatting
decisions themselves need to be taken at the appropriate level of
linguistic abstraction.

The finer
grained version of the NLG architecture proposed here is made of six
components: Text Planner, Sentence Planner, Sentence Grammar,
Morphological Synthesizer, Phonological Component, Formatting
Realiser. The corresponding representation levels are:
(1,Msg)\footnote{($<$level number$>$,$<$abbreviation$>$) are given for easying
  future reference.} Message, (2,TPlan) Text Plan, (3,SRep) Sentence
Representation, (4,MBundle) Morphological Bundle, (5,PWord) Potential Word,
(6,Str) String and (7,Frm) Formatting Instructions.\footnote{Formatting
  instructions can be introduced at any level, so we mention level
  (7,Frm) only as an abstraction.}

In a
hybrid architecture all levels can be mixed.  
 Mixing
representation levels can be done either by concatenation or by
embedding.  {\it Concatenation} means building a list of objects
pertaining to different levels. For instance, one can have a list with
the structure [$<$string$>$, $<$potential word$>$,$<$sentence
representation$>$,$<$string$>$]. Mixing by {\it embedding} means that
an object of a certain level is nested inside a structure of a
different level. Only structured objects can be the locus of an
embedding. Fixed strings and potential words can be
embedded, but one cannot embed into them.

Let us discuss a few cases from the literature in the light of the
theoretical framework that we are proposing for hybrid systems. The generation
techniques proposed in IDAS \citep{ReiterMellish93,
ReiterMellishLevine95} mix representations levels in two ways.
Knowledge base references to entities can be embedded into portions of
canned text, which gives a solution of the type [Msg, Str]. They also
fill case frame slots with canned text, which corresponds to a type
[SRep(Str)].

\cite{Busemann96} presents TG/2, a surface generator taking as input
formulae of the GIL sentence representation formalism. This system is
based on production rules employing canned text, templates and
syntactic representations. The production rules can contain calls to
other rules, lines of Lisp code and canned text. The whole proposal
seems to be a solution of the type [Msg, SRep, Str], where messagges
are picked up through direct Lisp calls. Then,
\cite{BusemannHoracek98} do away with the GIL representation interface
and replace it with an Intermediate Representation layer that is made
up of domain specific conceptual structures. More on this approach in section 4.2.

\cite{GeldofVelde97} propose a template based system for generating
hypertexts. They use templates made up of canned text interleaved with
``abstract terms referring to domain concepts''. This type of template
corresponds, in their words, to the IDAS' solution of type [Msg, Str]
mentioned above. Then, there are templates with hypertext
links. Finally, a text schema is used to structure the text. This
gives a solution of the type [TPlan([Msg, Str, Frm])].

\section{The Hyper Template Planning Language \label{HTPL}}

In  order    to extend the   potentiality  of   the  hybrid  approach,
\cite{Cancedda97} developed a  specific representation language, Hyper
Template  Planning Language (HTPL), which allows  one  to mix together
MBundle, PWord, Str  and Frm. 
We call {\em flexible} templates the kind of structures that can be
built by mixing these representation levels.  Recall that static
templates were defined above as operating on fixed strings.

Then,  in \citep{PiantaTovena98gen} the
expressive power  of HTPL has been extended  by adding the possibility
to mix  also Messages and Sentence Representations. Here is a list of
the linguistic representation  levels available in the current version
of HTPL. 

\paragraph{message representation (1,Msg):} a formula in some content 
representation formalism.  When  specifying  a message representation,
one should also  specify the formalism and the  type of message object
which is being  described.   For instance, {\tt  msg('IF',} {\tt attribute,}
  {\tt loca\-tion=pittsburgh)}  can be used   to refer to  an attribute-value
pair   of  the    Interchange  Format   (IF)   representation language
\citep{TovenaPianta99:iwcs}. Message representations are handled by a 
specialized component during the interpretation process. In the 
current version of HTPL, messages can by at most proposition 
level content specifications. Thus, the component that handles them is 
in fact a Sentence Planner.

\paragraph{phrase representation (3,SRep):} an abstract representation of a phrase 
which can feed a specific tactical generator. For instance, we
can specify phrases in terms of grammatical functions such as subject,
verb, object, adjuncts,  determiner etc., in the  spirit of LFG.  

\begin{verbatim}
phrase(lfg, 
  [subject=
   [spec=the,num=sing,pred=room]])}.
\end{verbatim}

\paragraph{morphological bundle (4,MBundle):} a set of morphological features 
corresponding to a word form. For example, the bundle {\tt 
morpho([cat=noun, pred=room, num= plur])} can be seen as an abstract 
representation of the word form {\tt rooms}. When used in HTPL expressions, 
the values of morphological features can be variables: {\tt 
mor\-pho([cat=noun, pred=room, num=} {\tt NUM])}.  Morphological variables 
make it easier to treat agreement phenomena, which are awkward to 
handle with static templates.

\paragraph{potential word (5,PWord):} a word form which can undergo
phonological adjustment. We describe a potential word by specifying the
lexical category of the word and its base form: {\tt w(noun,
  albergo)}. Sequences of potential words are mapped onto
strings by phonological and orthographic rules: for example in Italian 
{\tt [w(prep, di), w(article,il), w(noun, albergo)]} becomes 
{\tt ["dell'albergo"]}. The preposition {\tt di} is first combined 
with the article {\tt il} yielding the compound form {\tt del} (of the). 
The latter combines with a noun beginning with a vowel yielding a 
contracted word group which is orthographically represented as 
"dell'albergo" (of the hotel).

\paragraph{string (6,Str):} a sequence of characters inserted in the 
text without modification, for instance: "hotel reservation".

\medskip

The representation levels have been listed here following an ordering 
which is relevant for the HTPL interpreter, see below. However the ordering 
is also meaningful from a linguistic point of view. A phrase representation is 
linguistically less abstract than a message representation. A potential word 
pertains to a less complex constituency level than a phrase representation. 
A potential word is less abstract than a potential word, etc.

Objects of level 1
through 4 are all parametric, i.e. they can contain variables which
are instantiated at processing time. This allows the possibility of
sharing information between objects of different levels.

An HTPL expression can include any combination of the above 
representation levels. Both concatenation and embedding are possible. 
Here is an example of concatenation:

\begin{verbatim}
[w(pronoun, 'I'), w(modal, will), 
"arrive at", w(article, the), 
morpho([cat=noun, 
       pred=airport, 
       num=sing]),
msg('IF', attribute, time=sunday)]

\end{verbatim}

and here is an embedding:

\begin{verbatim}
phrase(lfg, 
  [subject=
     htpl([w(pronoun, 'I')]),
  modality=will 
  verb=
     htpl([
       "arrive at", 
       w(article, the), 
       morpho([cat=noun, 
              pred=airport,
              num=sing])]), 
  adjuncts=
     htpl([
       msg('IF', attribute, 
           time=sunday)])
   ])
\end{verbatim}

Both these HTPL expressions correspond to the sentence {\it I'll arrive 
at the airport on Sunday}. Of course the first expression can be realized more 
efficiently than the second, as it doesn't need to be handled by the 
Sententence Generator. However the second allows more flexibility; 
under certain conditions, the Sentence Generator could topicalize the 
adjunct yielding {\it ON SUNDAY, will I arrive at the airport!}. 
Also, note that what can be embedded is not simply canned text but 
any legal HTPL expression. Embedding is explictly marked by enclosing 
the embedded expression in the scope of the {\tt htpl} operator.

\paragraph{formatting (7,Frm):} Typographical formatting phenomena are handled in HTPL by 
including basic expressions in the scope of one ore more formatting 
operators such as: {\tt italic}, {\tt bold} etc. Hypertextual links are treated 
as a special class of format instructions. They are specified by 
descriptors which refer to linked documents through absolute 
addresses (file name) or functional expressions, evaluated at run time. 
Pictures are inserted in text through the same mechanism. Here follow 
other HTPL objects.

\paragraph{slot specifications:}{\tt slot(}$<$parameters$>${\tt)}. 
The run-time evaluation of a slot specification is expected to yield an HTPL expression.

\paragraph{template definitions:} {\tt template(}$<$template descriptor$>$, $<$HTPL 
expression$>$ {\tt)}. The $<$template descriptor$>$ can include variables, 
thus allowing the definition of parametric templates.

\paragraph{control expressions:} {\tt if\_then, if\_then\_else,}~ {\tt or}. 
In conditional expressions, an HTPL expression is realized in the generated text only if some constraint is satisfied. Here is an example of a
template definition including a conditional expression and a 
recursive call to other templates:

\begin{verbatim}
template(controls(ActID),
  if_then_else(
    exist_many_controls(ActID),
    template(item_controls(ActID)),
    template(coord_controls(ActID))))
\end{verbatim}

Disjunctive expressions give alternative ways of phrasing something, 
for example:  {\tt or(["taking} {\tt into} {\tt acc\-ount"}, {\tt "considering"])}. When 
the HTPL interpreter finds a disjunctive expression for the first 
time, it chooses one of the alternatives randomly; the second time, 
one of the remaining alternatives is selected, and so on. When all 
alternatives have been used at least once, the whole set becomes available again.

\subsection{The HTPL interpreter}
The HTPL interpreter must be able to handle both concatenated and embedded 
Mixed Representations (MR). As for concatenated MRs, the interpreter
scans the list of objects several times.
 The first time, it calls the component appropriate for all objects of
level (1, Msg) that is a Sentence Planner.  Then, it passes
all the objects of level (2, SRep) to the Sentence Grammar, and so
on up to objects of level (6, Str), which are passed to the
Formatting Realiser. The latter translates formatting instructions in 
HTML tags.  Note that each component related to a certain
level can produce as output MRs, although of a less abstract 
level.

Handling embedded MRs is more complex, as it depends highly on the
working of the single components. For this reason, solving embedded
MRs is the responsibility of each component, and won't be further
discussed here. The only generic constraint enforced by the HTPL
intepreter is that an object of level $n$ should not
embed objects of higher abstraction levels.

\subsection{Mixed Representations vs Intermediate Representations}
In section \ref{hybrid} we already analysed various hybrid approaches
to text generation in terms of the MRs framework. In this section we
will make some additional comparison between our proposal and that in
\citep{BusemannHoracek98}. The two approaches share many practical
motivations and adopt a number of similar or equivalent technical
solutions. There is one point however that sets well apart the two
approaches. \cite{BusemannHoracek98} introduce Intermediate
Representations, which can be characterized as language independent
but domain dependent representations. Notice that the domain
dependency holds not only at the level of the concepts of the ontology
but also at the level of the syntax and the intepretation of the
representation language. In other words, Intermediate Representations
are very different both from language dependent grammar representation
formalisms such as SPL, and from knowledge representation formalisms
based on a general syntax and a general
 semantic interpretation mechanism. As
the authors themselves suggest, the use of these kind of
representations seriously undermines the standard NLG architecture as
it doesn't acknowledge the text analysis levels on which that
architecture is based. We think that the notion of Mixed
Representation does not have the same reflexes on the NLG
architecture.  In our proposal, all analysis levels are kept, indeed
some more are made explicit. What is different with respect to the NLG
architecture is the possibility to introduce precompiled knowledge at
any stage of the generation process. This, from a processing point of
view, corresponds to the ability to skip unnecessary intermediate
processing stages.

\section{Conclusion}

In this paper we discussed the hybrid approach to automatic text
generation.  The concept of mixed linguistic representation turned out
to be a core notion for building a theoretical framework within which
to represent different attempts to combine NLG and template based
approaches. This conceptual framework led us to propose a more
detailed version of the standard NLG architecture and hence new types
of mixed representations.  These ideas were implemented in HTPL,
which has been successfully used in two applicative projects,
see \citep{Cancedda97, PiantaTovena98gen}.

\end{document}